\documentclass[letterpaper, 10 pt, conference]{ieeeconf}  

\IEEEoverridecommandlockouts                              

\usepackage{amsmath,amsfonts}
\usepackage{algorithm}
\usepackage[noend]{algpseudocode} 
\usepackage{array}
\usepackage[caption=false,font=normalsize,labelfont=sf,textfont=sf]{subfig}
\usepackage{textcomp}
\usepackage{stfloats}
\usepackage[hyphens]{url}
\usepackage{verbatim}
\usepackage{graphicx}
\usepackage{svg}
\usepackage{cite}
\usepackage{tabularx, booktabs}
\newcolumntype{C}{>{\centering\arraybackslash}X} 
\usepackage{siunitx}
\usepackage{multirow}

\newcommand{\footnoteurl}[1]{\footnote{\scriptsize\url{#1}}}
\newcommand{\captionurl}[1]{\scriptsize\url{#1}}

\title{\LARGE \bf
Adaptive Manipulation using Behavior Trees
}

\author{Jacques Cloete$^{1}$, Wolfgang Merkt$^{1}$, and Ioannis Havoutis$^{1}$
\thanks{$^{1}$The authors are with the Dynamic Robot Systems (DRS) group, Oxford Robotics Institute, University of Oxford, Oxford, UK
        {\tt\small \{jacques, wolfgang, ioannis\}@robots.ox.ac.uk}}%
}

\renewcommand{\baselinestretch}{0.95}

\begin{document}

\maketitle
\thispagestyle{empty}
\pagestyle{empty}


\begin{abstract}
Many manipulation tasks pose a challenge since they depend on non-visual environmental information that can only be determined after sustained physical interaction has already begun. This is particularly relevant for effort-sensitive, dynamics-dependent tasks such as tightening a valve. To perform these tasks safely and reliably, robots must be able to quickly adapt in response to unexpected changes during task execution, and should also learn from past experience to better inform future decisions. Humans can intuitively respond and adapt their manipulation strategy to suit such problems, but representing and implementing such behaviors for robots remains a challenge.
In this work we show how this can be achieved within the framework of behavior trees. We present the adaptive behavior tree, a scalable and generalizable behavior tree design that enables a robot to quickly adapt to and learn from both visual and non-visual observations during task execution, preempting task failure or switching to a different manipulation strategy.
The adaptive behavior tree selects the manipulation strategy that is predicted to optimize task performance, and learns from past experience to improve these predictions for future attempts.
We test our approach on a variety of tasks commonly found in industry; the adaptive behavior tree demonstrates safety, robustness (100\% success rate) and efficiency in task completion (up to 36\% task speedup from the baseline).
\end{abstract}


\section{Introduction}
\label{introduction}

Service robots can revolutionize how entire service-oriented domains work, with potential applications including autonomous maintenance in -- often remote -- industrial facilities \cite{Industry, Industry_2}, such as oil rigs \cite{OilRig}, or care robots in domestic environments \cite{DomesticCare_2, DomesticCare_3, DomesticCare}. To complete tasks effectively, robots must use observations from the environment to construct and execute a suitable plan; a number of frameworks have been developed for task and motion planning (TAMP) for manipulation \cite{TAMP_Survey, TAMP_Applications_Limitations, TAMP}.

This work is motivated by effort-sensitive, dynamics-dependent manipulation tasks for articulated devices commonly found in industrial settings, for example twisting a valve until tightened to a specified torque, pushing an e-stop button until pressed, or firmly picking and placing a fragile object. Robots generally require highly controlled environments with known task parameters for reliable planning and execution of manipulation tasks; however, the tasks we are interested in pose a significant challenge since they depend on information that can only be determined after sustained physical interaction has already begun. For example, a robot cannot determine the initial tightness of a valve or stiffness of a button from visual data alone, and can only gather this information by beginning to interact with it. Even then, in the absence of an accurate model for the dynamic profile for the device, the robot cannot know for how long it must twist the valve, or how hard it must push the button, to complete the task. Even if provided a model, stiction from corrosion or unclear servicing may lead to uncertainty and deviation from standard assumptions. To solve these kinds of tasks, we argue that the robot must be able to instead decide on an initial plan with which to start the task and adapt to observations it makes during task execution. Since the tasks we consider may require the robot to apply a significant amount of effort to manipulate the device, adaptability is not only critical for task completion but also for safe robot operation; failure to respond to a sudden unexpected spike in effort can cause catastrophic failure and even damage to the robot or environment.

Existing TAMP frameworks generate plans based on the current task environment, enabling robots to adapt to uncertain and changing environments; however, this adaptability is restricted to visually observable changes, such as for real-time collision avoidance. We require a solution that can handle unexpected deviations that are not visually observable.

\begin{figure}[!t]
\def\FigureScale{0.47}
\centering
\subfloat[]{\includegraphics[width=\FigureScale\linewidth]{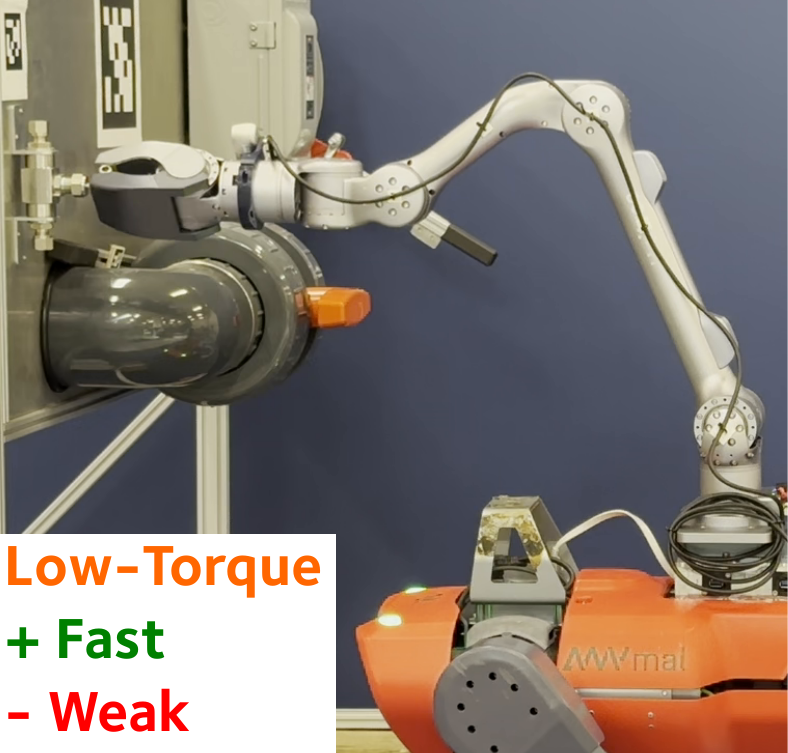}%
\label{fig:high_and_low_torque:low_torque}}
\hfil
\subfloat[]{\includegraphics[width=\FigureScale\linewidth]{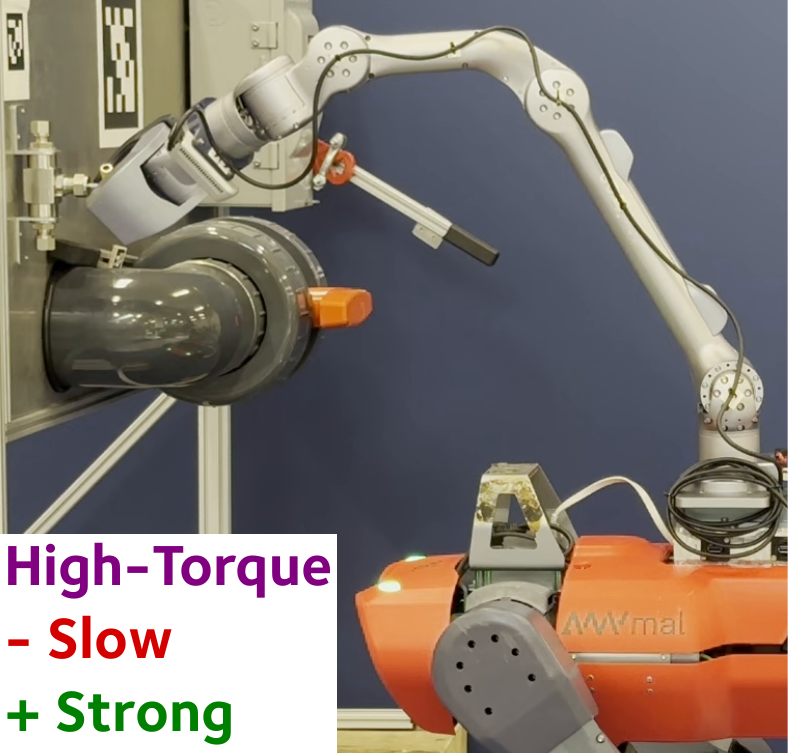}%
\label{fig:high_and_low_torque:high_torque}}
\caption{Different strategies for tightening a needle valve. 
(\ref{fig:high_and_low_torque:low_torque}) Quickly grasping head-on means that all applied torque is generated purely from the wrist joint effort, producing a fast but `low-torque' strategy. 
(\ref{fig:high_and_low_torque:high_torque}) Grasping from the side requires more complex motion but enables the robot to use the arm as a wrench, producing a slow but `high-torque' strategy.
}
\label{fig:high_and_low_torque}
\vspace{-8pt}
\end{figure}

We also consider the utility of a robot having access to different manipulation strategies for the same task. A \textit{strategy} is the specific approach that a robot may take to carry out a task. For manipulation tasks, different strategies may involve grasping the object differently, or even using a tool to assist manipulation. Different strategies can have different trade-offs, for example speed, complexity, effort and power expenditure; Fig. \ref{fig:high_and_low_torque} presents examples for tightening a valve.
We seek a system that can smartly select and switch between known strategies, ideally optimizing for performance by learning from past manipulation attempts.

We tackle both problems at the task planning level within the framework of behavior trees, exploiting their reactivity and modularity. The key contribution of our work is the adaptive behavior tree, which enables a robot to handle uncertainty in the environment and quickly adapt to unexpected safety-critical changes during task execution, preempting task failure or switching to a more optimal strategy based on data gathered over previous attempts. It comprises two stages; a selector that uses data recorded over previous attempts to predict the optimal strategy to deploy, and a monitor that checks a set of conditions concurrently to strategy execution, halting the current strategy if the task is complete or a safety condition is violated and re-attempting the task if necessary. The design is highly modular and generalizable, allowing for easy scaling to new strategies and tasks. We test the adaptive behavior tree on a diverse range of case studies representing complex, effort-intensive manipulation tasks from industry.



\section{Related Work}
\label{related_work}

\subsection{Adaptive Control for Robotics}

It may be argued that our problem of manipulating articulated devices with unknown and changing dynamics intrinsics can be solved with an appropriate adaptive controller for each task. Adaptive control has existed in the literature for many years, and generally focuses on online parameter estimation and learning for uncertain dynamic systems. There exist many good surveys on the literature for adaptive control \cite{AdaptiveControl, AdaptiveControl_2}. The approach has been extended to robotic manipulation tasks, for example estimating model parameters for an uncertain end-effector payload \cite{AdaptiveControl1, NonBTAdaptiveManipulation}.

However, we argue that adaptive behavior by parameter tuning is inherently limited in the diversity that it provides for adaptability, and does not readily allow for adaptability to the extent of reactively switching to a completely different manipulation strategy, such as grasping the device in a different way or using a tool, which may be necessary for task completion. Thus, the adaptive behavior we seek must go beyond the control level to the planning level; we opt to instead tackle the problem as a task planning problem. Doing so also allows us to define a more diverse range of conditions for task success and when to halt a strategy, and how to handle when these conditions are met. We will see that our approach can enable a robot to achieve impressive task performance despite only using out-the-box controllers.

\subsection{Behavior Trees for Robotics}

There is much work on task planning and autonomous decision-making for robotics, with the state of the art typically adopting a Finite State Machine (FSM), Behavior Tree (BT) or Partially Observable Markov Decision Process (POMDP) \cite{BT, BT_Intro}. The appropriate choice depends on the application and intended robot behavior.

A BT is a hierarchical directed tree that controls the order of execution of all the tasks that the robot may perform, each represented as a leaf node. Subtrees representing simpler behaviors can be used to build more complex ones, giving BTs a very composable and modular structure. During normal operation a `tick' signal is propagated between the root of the tree and its leaves, with the active leaf node either returning \textit{Success}, \textit{Failure} or \textit{Running}. This return signal determines which node is executed on the subsequent tick. It can be demonstrated that BTs offer far more modularity than FSMs \cite{BT_Intro, BTandFSM, BTvFSM}, and BTs also offer superior reactivity by allowing for the concurrent execution of different tree nodes that can interrupt each other if needed. BTs will be our planning framework of choice since the reactivity that they offer is critical for the manipulation tasks we consider, and our approach will also exploit their modularity.

There are many examples demonstrating how BTs can be used to coordinate manipulation tasks \cite{BT_Manipulation_Industrial, BT_Verbal, BT_FMEA} and robot control \cite{BT_Control, BT_Control_2, BT_SOT}; \textit{Iovino et. al.} provide a comprehensive survey of BTs for robotics \cite{BT}. BTs for robotics can be hand-coded \cite{BT_Hand_coded, BT_Hand_coded_2, BT_Hand_coded_3}, planned analytically \cite{RBTs, RBTs_2, BT_LTL, BT_PDDL, BT_RDL, BT_TAMP}, learned from demonstration \cite{BTsFromDemonstration, LegsAsManipulator, Jain2024} or even generated using Large Language Models (LLMs) \cite{Zhou2024, Ao2024}.

Reactive BTs are particularly relevant to our problem; they use a planning algorithm to automatically create and update a BT on the fly based on observations during task execution, achieving robot behavior that is adaptive to uncertain and changing environments \cite{RBTs, RBTs_2}. While this is very promising work, due to the safety-critical nature of the manipulation tasks we aim to tackle, we will opt for an approach that provides a more predictable method of handling failure preemption and recovery. 

On the other hand, \textit{Wu et. al.} \cite{BT_FMEA} propose an automated BT error recovery framework that uses a Bayesian Network (BN) as a decision model for error recovery upon task failure, alongside a human-made Failure Mode and Effects Analysis (FMEA) chart used to inform the network of possible sources of error as well as the corresponding solutions. In contrast, we desire a robot behavior that can preempt failure entirely by stopping and re-attempting the task once a safety condition is violated, potentially using a different strategy.

\section{Behavior Trees -- An Overview} 
\label{BTs_overview}

\begin{figure*}[!t]
\centering
\vspace{4pt}
\includegraphics[width=0.75\linewidth]{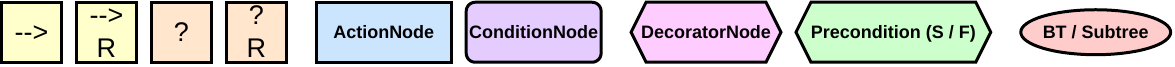}
\caption{Categories of BT tree nodes used in this work. From left to right; ControlNode (Sequence, ReactiveSequence, Fallback, ReactiveFallback), ExecutionNode (ActionNode, ConditionNode), DecoratorNode, Precondition, BT.}
\label{fig:BT_node_categories}
\end{figure*}

Without loss of generality, we follow the BT notation used by BT.CPP \cite{BT.CPP}; all BT nodes used in this work fall under one of a set of categories represented in Fig. \ref{fig:BT_node_categories}.

A \textbf{ControlNode} controls the sequence of execution of its children; those used in this work are of type Sequence, ReactiveSequence, Fallback or ReactiveFallback.

A \textbf{Sequence} propagates the tick to its children from left to right, returning \textit{Success} if and only if all their children return \textit{Success}. When it receives \textit{Running}, it returns \textit{Running} and ticks the same child on the next tick, and when it receives \textit{Failure}, it returns \textit{Failure} and resets the sequence progress. A \textit{ReactiveSequence} is similar, but when it receives \textit{Running}, it returns \textit{Running} and also ticks all children up to the current child on the next tick; thus all prior children are still checked for \textit{Failure}, enabling reactive behavior.

A \textbf{Fallback} (known as a Selector in other works) propagates the tick to its children from left to right until one of them returns \textit{Success}, at which point the Fallback stops and returns \textit{Success}. When it receives \textit{Running}, it returns \textit{Running} and ticks the same child on the next tick. It returns \textit{Failure} only if all children return \textit{Failure}. A \textit{ReactiveFallback} is similar, but when it receives \textit{Running}, it returns \textit{Running} and also ticks all children up to the current child on the next tick; thus all prior children are still checked for \textit{Success}, enabling reactive behavior (much like the ReactiveSequence).

An \textbf{ExecutionNode} carries out a command, forming the leaves of the BT, and can be of type ActionNode or ConditionNode. An \textit{ActionNode} performs a command, returning \textit{Success} if the command is successfully completed, \textit{Failure} if failed, or \textit{Running} if ongoing. A \textit{ConditionNode} checks a condition, returning \textit{Success} if true or \textit{Failure} if false.

A \textbf{DecoratorNode} modifies the return status of a single child. A \textit{Precondition} is a special type of DecoratorNode that ticks the child only if a precondition is met, or else returns either \textit{Success} or \textit{Failure} (marked by `S' or `F' respectively).

\section{Adaptive BT}
\label{proposed_adaptive_BT_design}

\begin{figure}[!t]
\centering
\includegraphics[width=0.995\linewidth]{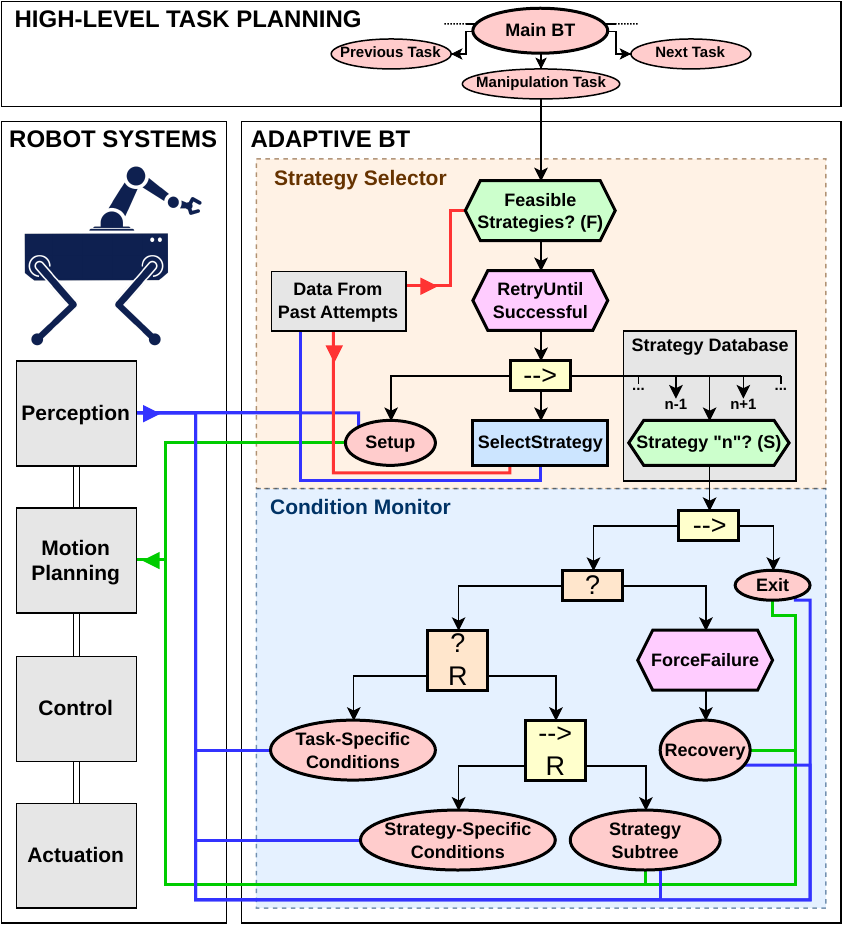}
\caption{The adaptive BT, integrated within a wider robot system representing the test platform in this work. The colored lines represent communication lines for perception (blue), motion planning (green) and strategy selection (red).}
\label{fig:adaptive_BT}
\vspace{-8pt}
\end{figure}

Fig. \ref{fig:adaptive_BT} presents the adaptive BT, as well as a demonstration of how it can be integrated within a wider robot system. The adaptive BT is composed of two core stages; the strategy selector first decides which strategy to use for the next attempt, based on current perception data as well as data recorded over past attempts (such as joint efforts or time taken). The condition monitor then wraps around the subtree for the selected strategy, checking a set of task- and strategy-specific conditions concurrently to strategy execution. The adaptive BT also admits subtrees to coordinate robot setup prior to strategy selection, and for safely exiting or recovering from the current attempt in preparation for the next.

The database of the subtrees for all $N \in \mathbb{N}$ known strategies are inserted as modules within a Sequence, but all are preconditioned to only be executed if that specific strategy has been selected. Thus, if strategy $n$ is selected, only that strategy's subtree will be ticked, and all the other subtrees trivially return \textit{Success}. If no strategies are considered feasible for the current task instance, SelectStrategy returns \textit{Failure} and the adaptive BT exits as failed.

The condition checking is such that if the subtree of task-specific conditions returns \textit{Success}, the task is considered to have been completed successfully, and if it returns \textit{Failure}, the task is still ongoing. This follows the BT design principle of backward chaining \cite{BT_Intro}, commanding the robot to keep executing the strategy until task completion conditions are met. Meanwhile, if the subtree of strategy-specific conditions returns \textit{Failure}, a constraint has been violated during execution of the current strategy and the robot must halt the current attempt and try again, potentially selecting a different strategy. If it returns \textit{Success}, no violations have occurred. This logic makes the condition monitor reactive and interpretable. If there are multiple task- or strategy-specific conditions to check, they should be placed in a ReactiveSequence to ensure proper operation of the adaptive BT. Each condition is ticked for every tick of the strategy subtree, allowing the robot to rapidly respond to new observations; thus the inherent reactivity of BTs is exploited. If a condition is not relevant to the current stage of the strategy subtree, it can be skipped by wrapping with an appropriate Precondition.

The subtrees for exit and recovery allow the robot to safely exit the current attempt, due to task success or preempting failure respectively. A \textit{ForceFailure} is wrapped around the recovery subtree to signal that the task is not yet complete and a re-attempt is required.

Very little has been assumed concerning the algorithm behind how the next strategy is selected, or the implementation of any of the subtrees within the strategy selector or condition monitor. The adaptive BT is thus highly modular, and generalizes to a wide range of different problems and can satisfy all manner of tasks, specifications and constraints; all that is required are the subtrees for the task and strategy.

The adaptive BT can easily be added as a subtree within a greater BT structure, and is thus readily applicable to existing BT-based robot task planning systems  without major changes. The only required knowledge for adding the proposed design into a larger BT is that the adaptive BT must be repeatedly ticked while \textit{Running}, and will return \textit{Success} if the task was completed successfully and \textit{Failure} otherwise.

The adaptive BT is designed to coordinate efficient selection, monitoring and switching of strategies, not the synthesis of the strategies themselves. The adaptive BT would synergize well with methods for the automatic synthesis of BTs \cite{RBTs, RBTs_2, BT_LTL, BT_PDDL, BT_RDL, BT_TAMP,BTsFromDemonstration, LegsAsManipulator,Jain2024,Zhou2024,Ao2024} to generate new strategies for it to use, but exploring this is beyond the scope of this work.

\section{Adaptive BT for Manipulation}
\label{adaptive_BT_for_manipulation}

\begin{figure*}[!t]
\def\FigureScale{0.15}
\centering
\subfloat[]{\includegraphics[height=\FigureScale\textheight]{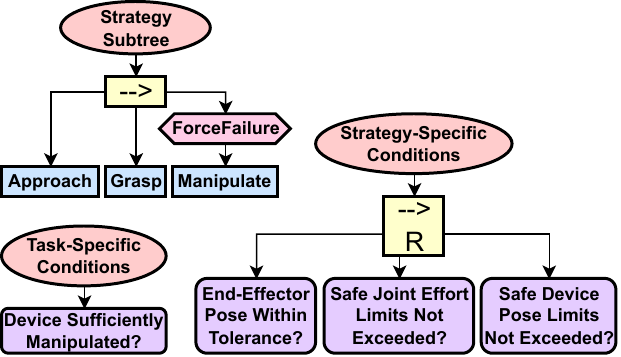}%
\label{fig:manipulation_BTs:manipulate_device}}
\hfill
\subfloat[]{\includegraphics[height=\FigureScale\textheight]{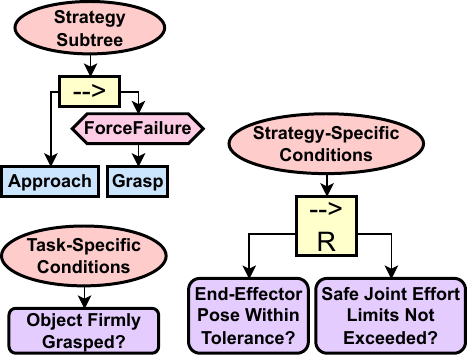}%
\label{fig:manipulation_BTs:grasp_object}}
\hfill
\subfloat[]{\includegraphics[height=\FigureScale\textheight]{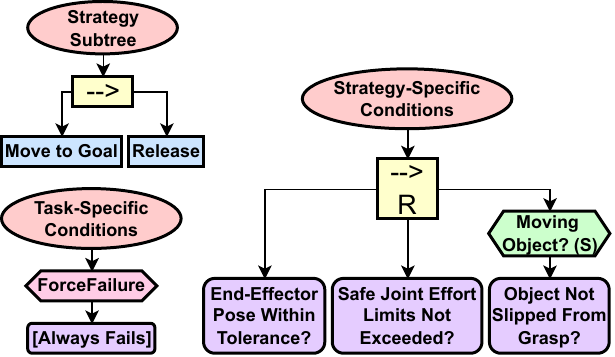}%
\label{fig:manipulation_BTs:move_object}}
\caption{Subtrees for strategies used in this work. 
(\ref{fig:manipulation_BTs:manipulate_device}) Manipulating an articulated device. 
(\ref{fig:manipulation_BTs:grasp_object}) Grasping an object.
(\ref{fig:manipulation_BTs:move_object}) Moving an object to a goal.
}
\label{fig:manipulation_BTs}
\vspace{-8pt}
\end{figure*}

We now apply the adaptive BT to manipulation tasks; Fig. \ref{fig:manipulation_BTs} presents the general subtrees for all strategies used in this work. We stress that these designs are just examples used to test the adaptive BT, with an emphasis on simplicity; the adaptive BT readily accepts new strategies for new tasks.


\textbf{Manipulating Articulated Devices.} Fig. \ref{fig:manipulation_BTs:manipulate_device} presents general subtree structures for strategies that manipulate an articulated device, where task success occurs when the device is considered sufficiently manipulated (e.g. valve is tight). The strategies consist of approaching, grasping and manipulating the device (e.g. twist by one-half rotation), while checking safety conditions. Note that these same subtree structures are used for both the low-torque and high-torque strategies depicted in Fig. \ref{fig:high_and_low_torque}, and even the strategies for manipulating the different devices (e.g. tightening a valve vs pressing a button), though the action and condition nodes have different parameters (or even implementations) for each strategy.

\textbf{Object Grasping.} Fig. \ref{fig:manipulation_BTs:grasp_object} presents general subtree structures for strategies that grasp an object, where task success occurs when the object is considered firmly grasped (e.g. by checking gripper effort). The strategies consist of approaching and grasping, while checking safety conditions.

\textbf{Object Placing.} Fig. \ref{fig:manipulation_BTs:move_object} presents general subtree structures for strategies that move an (already grasped) object to a goal pose, where task success occurs if the strategy completes without preemption (thus no task-specific conditions). The strategies consist of moving and then releasing the object, while checking safety conditions.

\textbf{Setup/Recovery/Exit.} In this work, these are simply action nodes that position the robot to view the scene for strategy selection (for setup) or reset the robot pose (otherwise).

\textbf{Strategy Selection.} The SelectStrategy node estimates the optimal manipulation strategy to use; for this work, the optimal strategy $a^{*}$ for task state $s$ minimizes task duration $t_{\mathrm{est}}$ while not violating strategy-specific constraints (often corresponding to the safety conditions, e.g. maximum safe joint efforts). This is equivalent to the following problem:
\begin{equation*}
    a^{*}(s) = \min_{a \in \mathcal{A}_{\mathrm{task}}} t_{\mathrm{est}}(a,s) \quad \text{subject to} \quad a \in \mathcal{A}_{\mathrm{safe}}(s),
\end{equation*}
where strategies $a$ are selected from the set of available task-relevant strategies $\mathcal{A}_{\mathrm{task}}$, and $\mathcal{A}_{\mathrm{safe}} \subseteq \mathcal{A}_{\mathrm{task}}$ is the set of feasible strategies given the constraints. Both $t_{\mathrm{est}}$ and $\mathcal{A}_{\mathrm{safe}}$ are estimated from data recorded from previous attempts; if no data exists at $s$, we let $\mathcal{A}_{\mathrm{safe}} = \mathcal{A}_{\mathrm{task}}$, and calculate $t_{\mathrm{est}}$ from nominal durations of the strategy's motion trajectories.

\section{Implementation}
\label{implementation}

\textbf{Test Platform.} We used an ANYbotics ANYmal quadruped
with mounted Unitree Z1 robotic arm
, equipped with a wrist-mounted Intel Realsense Depth Camera D345
for perception. ROS, the Robot Operating System, was used for the implementation of the test setup as well as communication with hardware \cite{ROS}.

\textbf{Behavior Tree.} BT.CPP 4.0 \cite{BT.CPP} was used to construct the adaptive BT. The ActionNode types specific to this work take the form of ROS action and service clients, implemented as derived classes of the \textit{StatefulActionNode} from the BT.CPP core library; this node type supports asynchronous actions and is recommended for this use case.



\textbf{Motion Planning.} This work uses a whole-body motion planner made using EXOTica, the EXtensible Optimisation Toolset \cite{EXOTica}. The planner solves an EXOTica \textit{DynamicTimeIndexedShootingProblem}, tracking pre-defined waypointed end-effector pose trajectories for the approach, grasp, manipulate and retract stages of the manipulation task for each strategy. The pose trajectories are defined relative to the pose of the object or device handle; in this way, the planner generalizes to arbitrary poses and handle angles (so long as robot joint limits permit the motion). In the case of planning failure, the planner action nodes return \textit{Failure}, prompting the adaptive BT to retry the task.

\textbf{Control.} This work uses the standard controllers provided with the ANYmal and Unitree Z1. The combined controller is by no means perfect, since the ANYmal controller has no awareness of the arm and cannot compensate for the induced wrench on the quadruped when the arm is outstretched. In practice, there was often non-negligible deviation between the true end-effector pose and that in the motion plan, especially when the arm was outstretched; to ensure safe operation, this deviation is checked by the condition monitor during manipulation.
Using a bespoke whole-body controller that compensates for both the motion of the quadruped and arm would improve performance, but this is beyond the scope of this work; instead, we demonstrate how our adaptive BT can be used to overcome such problems at the task planning level by autonomously preempting failure and retrying.

\textbf{Perception.} Apriltags \cite{apriltag} were used to identify the pose of objects as well as device handles in an `unrotated' state, and invariant template matching \cite{InvariantTemplateMatcher} was used to identify handle angles and the pose of the rotated handles.

\section{Case Studies}
\label{experiments}

All device manipulation tasks use the general subtree structures depicted by Fig. \ref{fig:manipulation_BTs:manipulate_device}, albeit with task- and strategy-specific implementations of the action and condition nodes. Appendix \ref{appendix:feasible_set_of_strategies} describes estimation of $s$ and $\mathcal{A}_{\mathrm{safe}}$ for all tasks.

\textbf{Needle Valve.} The robot twists a Tameson NLS-012 needle valve from unknown starting tightness until tight, and loosens it to a user-specified angle, with access to the two strategies depicted by Fig. \ref{fig:high_and_low_torque}. To demonstrate how the adaptive BT can be readily placed within larger BT structures, we also consider a walk-and-manipulate task where the robot navigates to, tightens, and navigates away from the needle valve (with access to both manipulation strategies, but the required applied torque for tightness reduced so as to be achievable by the `low-torque' strategy). The adaptive BT is inserted as a subtree within a larger BT that also commands navigation to and from the device (using the standard ANYmal navigation and locomotion software).

\begin{figure*}[!t]
\def\FigureScale{0.245}
\centering
\subfloat[]{\includegraphics[width=\FigureScale\linewidth]{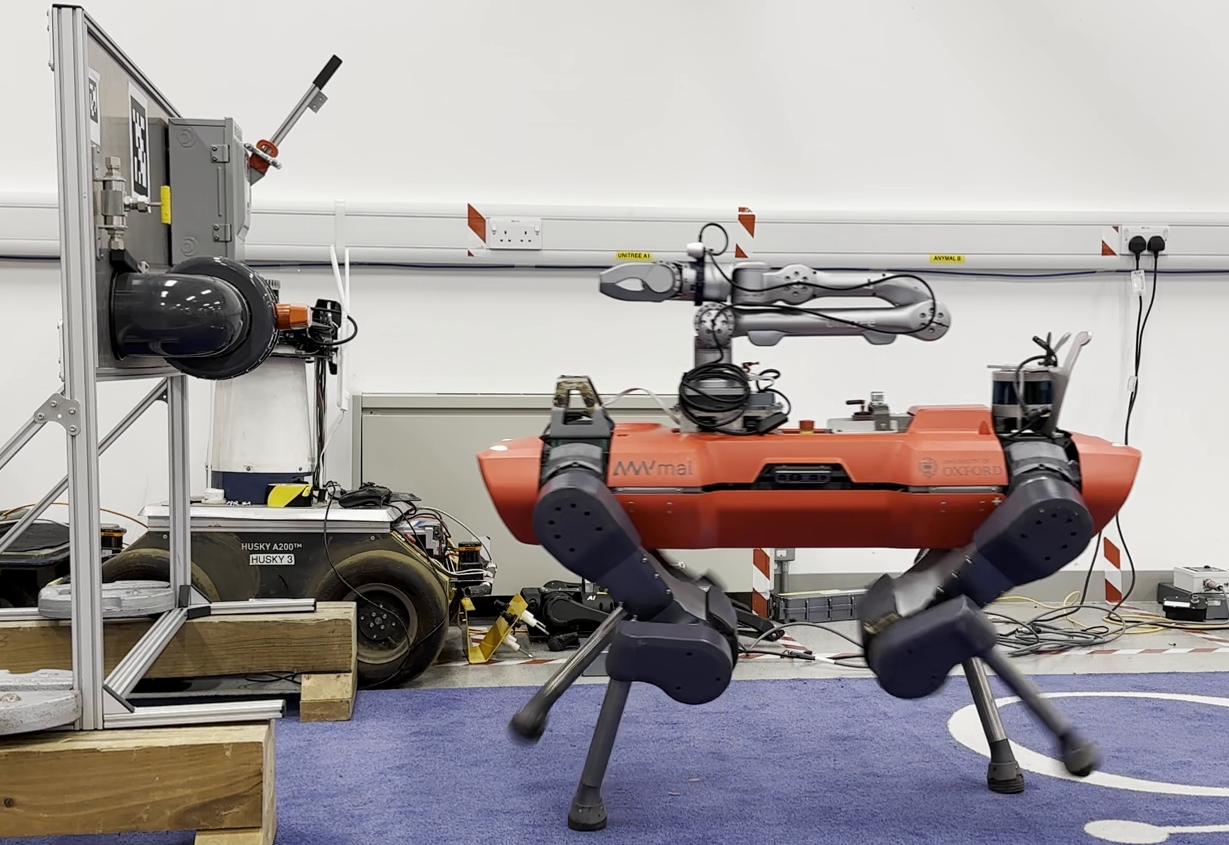}%
\label{fig:experiments:walk_to_needle_valve}}
\hfil
\subfloat[]{\includegraphics[width=\FigureScale\linewidth]{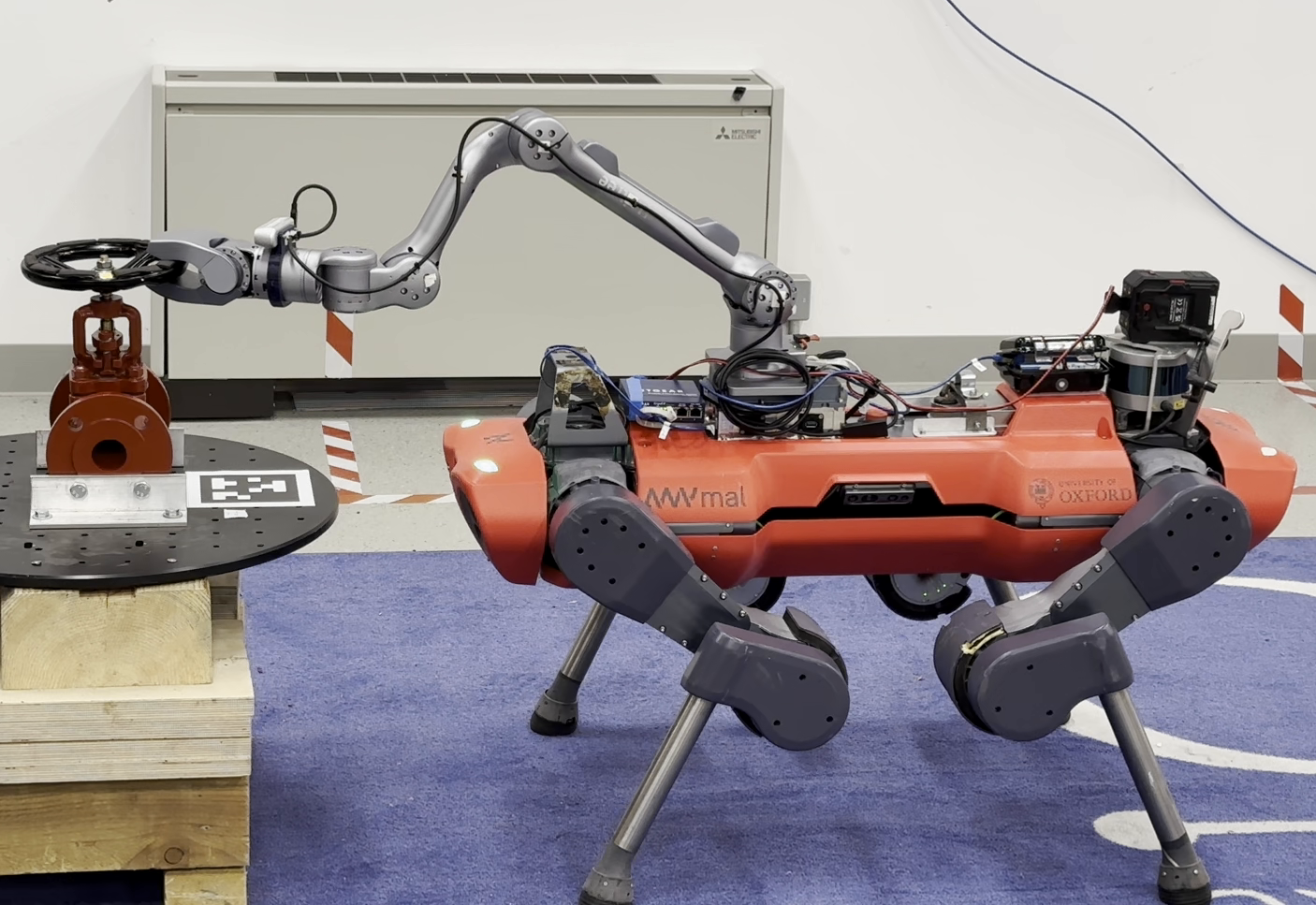}%
\label{fig:experiments:globe_valve}}
\hfil
\subfloat[]{\includegraphics[width=\FigureScale\linewidth]{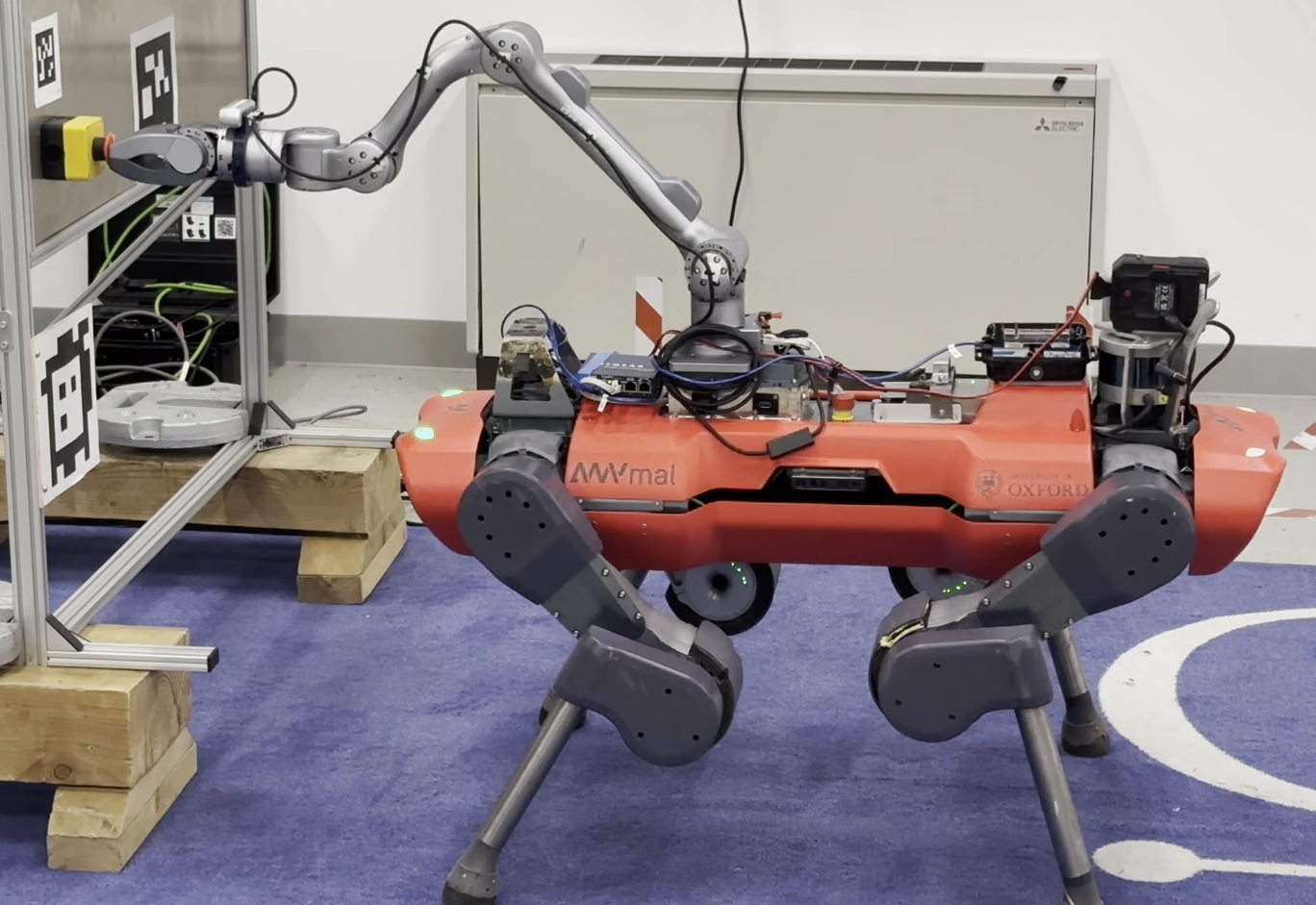}%
\label{fig:experiments:button_press}}
\hfil
\subfloat[]{\includegraphics[width=\FigureScale\linewidth]{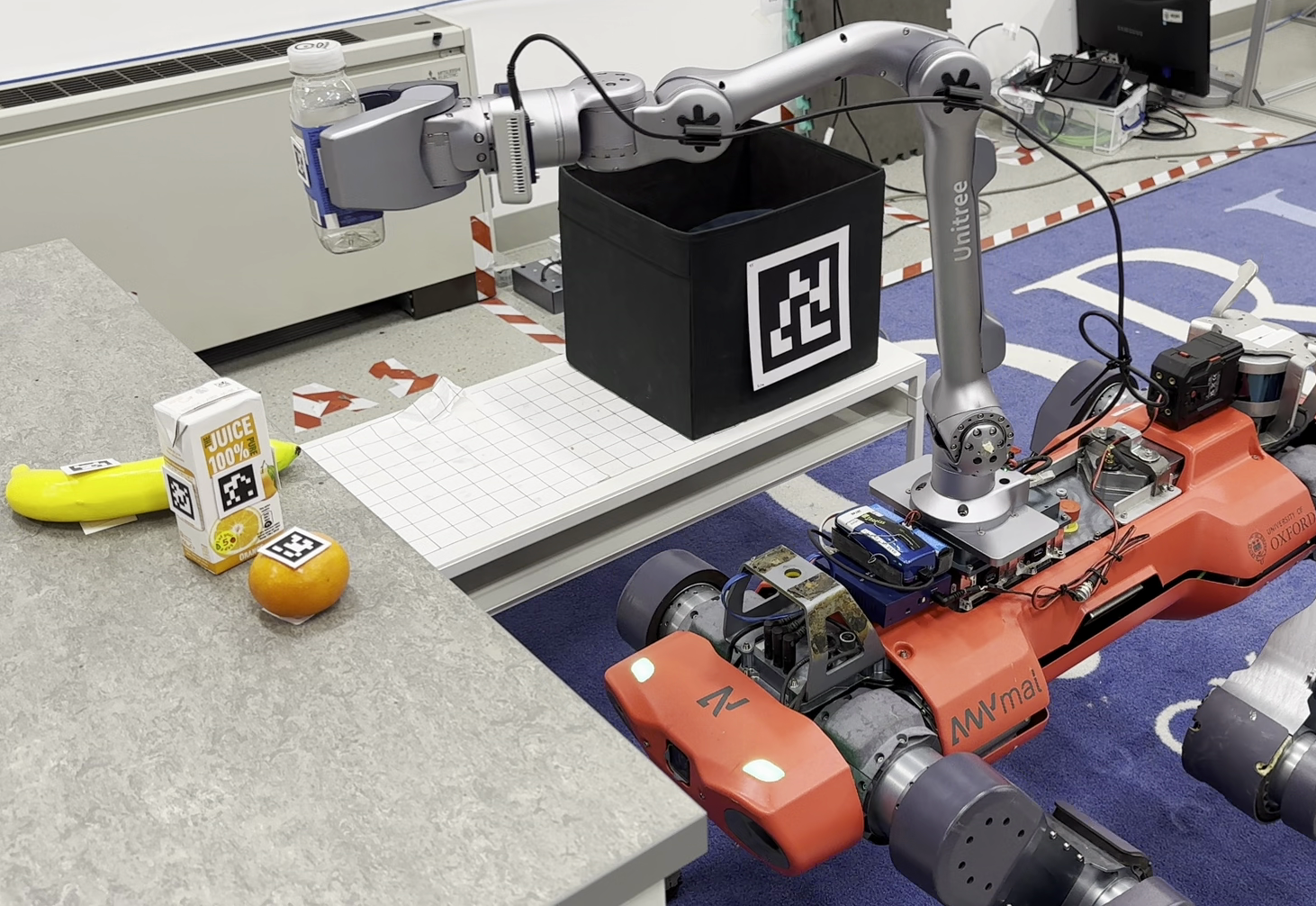}%
\label{fig:experiments:groceries_packing}}
\caption{Snapshots from experiments. 
(\ref{fig:experiments:walk_to_needle_valve}) Walking to the needle valve (as part of the walk-and-manipulate task).
(\ref{fig:experiments:globe_valve}) Tightening the globe valve.
(\ref{fig:experiments:button_press}) Pushing the e-stop button until fully pressed. 
(\ref{fig:experiments:groceries_packing}) Groceries packing.
Accompanying video with experiments: \protect\captionurl{https://youtu.be/im7OCkEiQsU}.
} 
\label{fig:experiments}
\vspace{-8pt}
\end{figure*}

\textbf{Globe Valve.} The robot twists a Dikkan DN40 PN16 globe valve from unknown starting tightness until tight, with access to a single strategy depicted by Fig. \ref{fig:experiments:globe_valve}.

\textbf{E-Stop Button.} The robot manipulates an IDEC HW1X-BV411-R emergency stop; it pushes the button until fully pressed, with access to a single strategy depicted by Fig. \ref{fig:experiments:button_press}, and twists it until released, with access to a single strategy very similar to that depicted by Fig. \ref{fig:high_and_low_torque:low_torque}.

\textbf{Groceries Packing.} The robot packs a plastic bottle, (full) juice box, (real) orange and (plastic) banana into a basket, in a user-specified order, as depicted by Fig. \ref{fig:experiments:groceries_packing}. The grasping stage uses a single strategy, with subtree structures depicted by Fig. \ref{fig:manipulation_BTs:grasp_object}, determining firm grasping by checking gripper joint effort. The packing stage has access to a continuous set of strategies; all strategies have the same subtree structures depicted by Fig. \ref{fig:manipulation_BTs:move_object}, but the speed of the motion trajectories is parameterized by a speed factor (i.e. a strategy) $a \in \mathcal{A}_{\mathrm{safe}}$, where $\mathcal{A}_{\mathrm{safe}}$ is determined by the weight of the object (lower maximum safe speeds for heavier objects, which may slip out of grasp when the motions are too fast). The adaptive BTs for the two stages are placed in sequence such that the robot returns to the first stage if the second stage is preempted.

\textbf{Baseline.} For all tasks, we use backward chaining with stacked Fallbacks as the baseline BT structure \cite{BT_Intro}. This is the standard BT design used when the robot must choose between different actions with robustness to failure. For fair comparison to the adaptive BT, we manually arranged the (nominal) `faster' strategies before `slower' ones in the baseline's Fallbacks such that the robot tries them first, with a reminder that the adaptive BT automates (and optimizes) the strategy selection process. Since the baseline cannot select from a continuous set of strategies, it can only (safely) use the lowest speed factor when grocery packing.

\section{Results and Discussion}
\label{results}

To evaluate the adaptive BT, we performed a series of ablations to assess how each component impacts performance, with comparison to the baseline where relevant. For our discussion, we refer to the results in Table \ref{table:AdaptiveManipulation}, which records, for 10 trials of each experiment, the cumulative success rate after each attempt (ignoring re-attempts triggered to change strategy and re-grasping when device handle pose limits were exceeded) and time taken for task completion. A maximum of 5 attempts was permitted for each trial, task success was verified by checking the scene after the robot had exited the task, and the initial device states were reset to be the same for each trial. Strategy effort limits were set to the Z1's maximum safe operating efforts scaled down by a safety factor, except for the needle valve low-torque strategy, which used a larger safety factor to avoid the quadruped becoming unstable from induced wrench during heavy twisting.

\begin{table*}
    \vspace{4pt}
    \caption{Experiment Results for All Tasks}
    \label{table:AdaptiveManipulation}
    \vspace{-10pt}
    \begin{center}
    \begin{tabularx}{\textwidth}{@{} l l *{5}{C} c c @{}}
        \toprule
        \multirow{2}{*}{Task} & \multirow{2}{*}{Experiment} & \multicolumn{5}{c}{Cumulative Success Rate (\%) After Attempt Number} & \multicolumn{2}{c}{Time Taken (s)} \\
        \cmidrule(lr){3-7}
        \cmidrule(lr){8-9}
         &  & 1 & 2 & 3 & 4 & 5 & Minimum & Mean \\
        \midrule
        \multicolumn{9}{l}{\textit{Ablation A: RFP Enabled (SAT + LPE Disabled, i.e. Equivalent to Baseline)}} \\
        \midrule
        \multirow{3}{*}{Needle Valve (Tighten)}
         & Low-Torque Strategy Only & 0 & Aborted & Aborted & Aborted & Aborted & 135 (Fail) & 136 (Fail) \\
         & High-Torque Strategy Only & 50 & 70 & 90 & 100 & Done & 279 & 336 \\
         & Walk-and-Manipulate & 60 & 80 & 90 & 100 & Done & 212 & 230 \\
        \midrule
        \multirow{2}{*}{Needle Valve (Loosen)}
         & Low-Torque Strategy Only & 0 & Aborted & Aborted & Aborted & Aborted & 58 (Fail) & 58 (Fail) \\
         & High-Torque Strategy Only & 70 & 90 & 100 & Done & Done & 228 & 247 \\
        \midrule
        Globe Valve
         & Tighten & 70 & 100 & Done & Done & Done & 95 & 107 \\
        \midrule
        \multirow{2}{*}{E-Stop Button}
         & Press & 80 & 100 & Done & Done & Done & 28 & 32 \\
         & Release & 70 & 80 & 100 & Done & Done & 45 & 63 \\
        \midrule
        \multirow{1}{*}{Groceries Packing}
         & Lowest Speed Only & 60 & 70 & 90 & 100 & Done & 261 & 269 \\
        \midrule
        \multicolumn{9}{l}{\textit{Ablation B: RFP + SAT Enabled (LPE Disabled)}} \\
        \midrule
        \multirow{2}{*}{Needle Valve (Tighten)}
         & Both Strategies (Baseline) & 70 & 100 & Done & Done & Done & 175 & 186 \\
         & Both Strategies (Adaptive BT) & 80 & 100 & Done & Done & Done & 175 & 182 \\
        \midrule
        \multirow{2}{*}{Needle Valve (Loosen)}
         & Both Strategies (Baseline) & 60 & 90 & 100 & Done & Done & 259 & 283 \\
         & Both Strategies (Adaptive BT) & 80 & 100 & Done & Done & Done & \textbf{204} & \textbf{212} \\
        \midrule
        \multirow{2}{*}{Groceries Packing}
         & Lowest Speed Only (Baseline) & 60 & 70 & 90 & 100 & Done & 261 & 269 \\
         & Weight-Based Speed (Adaptive BT) & 50 & 70 & 80 & 100 & Done & \textbf{196} & \textbf{206} \\
        \midrule
        \multicolumn{9}{l}{\textit{Ablation C: RFP + SAT + LPE Enabled (Complete Adaptive BT)}} \\
        \midrule
        \multirow{2}{*}{Needle Valve (Tighten)}
         & Both Strategies (Baseline) & 80 & 100 & Done & Done & Done & 94 & 126 \\
         & Both Strategies (Adaptive BT) & 90 & 90 & 100 & Done & Done & \textbf{58} & \textbf{118} \\
        \midrule
        \multirow{2}{*}{Needle Valve (Loosen)}
         & Both Strategies (Baseline) & 60 & 90 & 100 & Done & Done & 259 & 283 \\
         & Both Strategies (Adaptive BT) & 80 & 100 & Done & Done & Done & \textbf{172} & \textbf{182} \\
        \bottomrule
    \end{tabularx}
    \end{center}
\vspace{-14pt}
\end{table*}

\textbf{Reactive Failure Preemption (RFP).} For Ablation A, the robot only has access to one strategy (preventing strategy adaptation) and manipulation data is reset between trials (preventing learning from past experience). We see a 100\% success rate within the permitted number of attempts for all but one experiment when using the adaptive BT, despite having no knowledge of the valves' initial tightness, or how much they needed to be twisted to be tightened. The robot never suffered catastrophic failure since the condition monitor preempted failure before it occurred. When using the low-torque strategy, the robot aborted the needle valve tasks after exceeding the safe effort limit for the arm's wrist joint. This demonstrates the safety enabled by the adaptive BT, since the robot was able to react and safely exit the task before overexerting itself. The robot was also able to safely twist the valves for an arbitrary number of rotations without the danger of overextending itself, by periodically re-grasping the valves using the condition of exceeding safe device pose limits; once triggered, the robot re-attempted the task and re-grasped the device in a more favorable joint configuration for further twisting. The robot was also able to firmly grasp the different-shaped fragile objects without damaging them by reacting to spikes in gripper effort.


Under this ablation, the adaptive BT produces equivalent behaviors to the baseline. However, the adaptive BT enables easier scalability of task composition over the baseline, since the adaptive BT provides an abstracted, modular interface for appending and removing subtrees for strategies and conditions to the strategy database (see Fig. \ref{fig:adaptive_BT}), while editing the baseline backward chained BT structure can become unwieldy as the number of strategies and conditions grows.

\textbf{Strategy Adaptation (SAT).} For Ablation B, the robot has access to all strategies but manipulation data is still reset between trials (preventing learning from past experience). We see substantial average task speedups for needle valve tightening (46\%) and loosening (14\%) as well as groceries packing (23\%) when using the adaptive BT compared to Ablation A; the cause for this speedup is that the adaptive BT selects faster strategies when feasible and switches to slower, `safer' strategies when necessary.

\begin{figure*}[!t]
\centering
\vspace{4pt}
\includegraphics[width=0.99\linewidth]{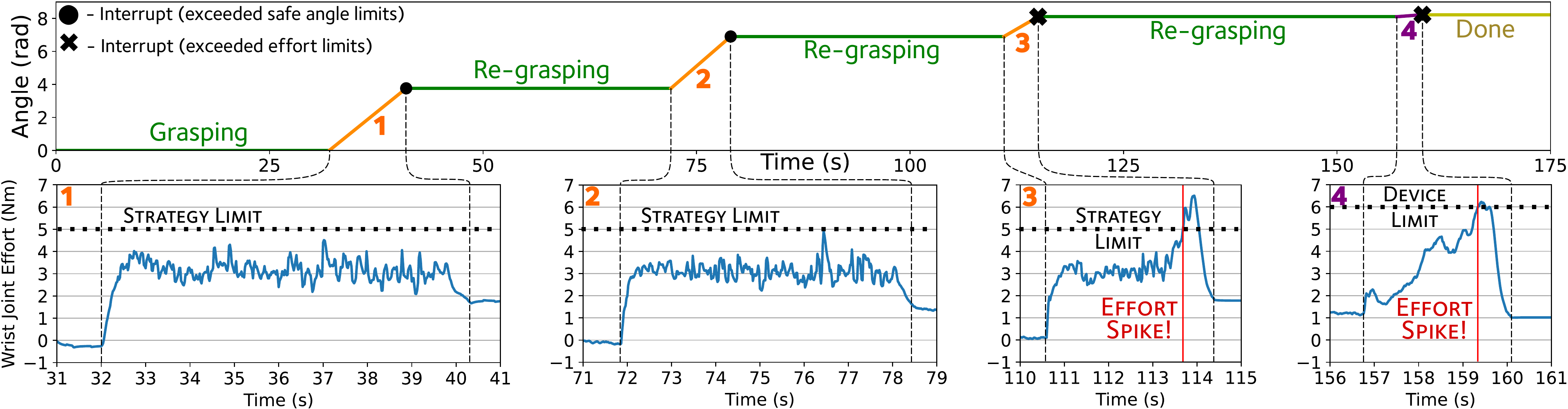}
\vspace{-8pt}
\caption{An annotated timeline of tightening the needle valve using the adaptive BT with both available strategies, with plots of the effort in the wrist joint during each manipulation attempt A, B, C (which all use the `low-torque' strategy) and D (which uses the `high-torque' strategy). The adaptive BT initially selects the low-torque strategy to twist quickly while the valve is loose but switches to the high-torque strategy near the end to apply more wrench and finish tightening. Note the effort spikes during attempts C and D; the first indicates that the low-torque strategy is no longer feasible, while the second indicates that the device is sufficiently tightened. For clarity, only the lowest active effort limit is shown in each plot.}
\label{fig:AdaptiveManipulationTimeline}
\vspace{-8pt}
\end{figure*}

For needle valve tightening, since the low-torque strategy is approximately 5 times faster while twisting, it is optimal for the robot to use it for most of the manipulation sequence, but when the robot experiences a sudden effort spike as the valve begins to tighten, it quickly adapts and uses the high-torque strategy to fully tighten the valve; Fig. \ref{fig:AdaptiveManipulationTimeline} provides an annotated timeline. We note that this is similar to behavior displayed by humans, twisting quickly when the valve is loose but changing grip near the end to be able to apply more wrench without hurting oneself. For this particular task the baseline produces the same behavior, leading to roughly the same average time taken; this is because the optimal behavior for this task (start low-torque, switch to high-torque at the end) can also be achieved using a simple Fallback. However in general this is not the case, as demonstrated by the needle valve loosening task, wherein the adaptive BT achieves a 25\% average speedup over the baseline. For this task, both the adaptive BT and baseline first (unsuccessfully) try the low-torque strategy (which cannot safely apply enough torque to loosen the valve) and then switch to the high-torque strategy for the first successful twist, after which the valve is loose enough to be loosened further using the low-torque strategy. While the logic of the Fallback means that the baseline will keep using the high-torque strategy for the remainder of the task, the adaptive BT optimizes over feasible strategies at each re-attempt, and thus chooses the (faster) low-torque strategy for the remainder of the task.

\textbf{Learning from Past Experience (LPE).} For Ablation C, the robot has access to all strategies as well as the manipulation data recorded from Ablations A and B; note that the baseline has no mechanism for learning from past manipulation data. For needle valve loosening, the adaptive BT achieves an even greater average speedup (36\%) over the baseline; this is because past manipulation data suggests that the valve will initially be too tight to be loosened with the low-torque strategy, so the robot has learned to start with the high-torque strategy (while in Ablation B, due to lack of knowledge of the valve's initial tightness, it would initially try low-torque then immediately switch to high-torque).

\begin{figure}[!t]
    \includegraphics[width=0.985\linewidth]{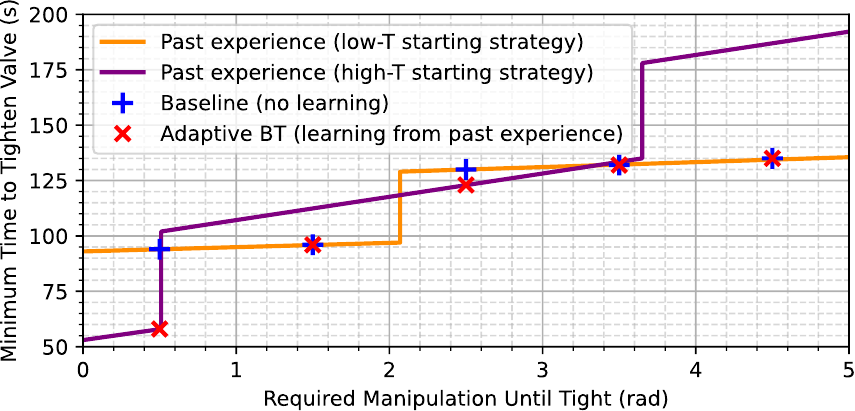}
\caption{Minimum time to tighten the needle valve for different (known) required rotations until tight. The adaptive BT starts with the strategy that minimizes time taken to tighten the valve according to past experience.}
\label{fig:learning_starting_strategy}
\vspace{-8pt}
\end{figure}

For needle valve tightening, we see no change from Ablation B if the robot has no initial estimate of the required rotation until tight, i.e. task state $s$. Note that as we vary initial $s$, in some cases it is faster to start with the high-torque strategy, since it avoids the overhead of switching from low-torque to high-torque to finish tightening. If we now provide the adaptive BT with an accurate initial estimate for $s$, it always starts with the strategy that minimizes time taken to tighten the valve according to past experience (unlike Ablation B, which always starts with low-torque). This is illustrated by Fig. \ref{fig:learning_starting_strategy}, which plots minimum time taken to tighten the needle valve for 5 different known required rotations, with the adaptive BT achieving up to 34\% speedup over the baseline (at 0.5 radians). For Ablation C's results for this task in Table \ref{table:AdaptiveManipulation}, the 10 trials are instead 2 trials each over these 5 different known required rotations; here the adaptive BT achieves 6.3\% average speedup over the baseline.

In practice, initial $s$ could be a random variable with a known distribution; for this case, the mean time taken over this distribution would be the appropriate $t_{\mathrm{est}}$ to use for strategy selection. If the distribution is unknown, it could be estimated empirically over many manipulations.

A limitation of the test setup is that the effort conditions for task success were found empirically from preliminary testing; however, we stress that this is unrelated to the adaptive BT itself, which assumes very little about each condition and would readily accept more sophisticated conditions based on task-space wrench estimation.

\section{Conclusion and Future Work}
\label{conclusion}

In this work we proposed the adaptive BT, a flexible and highly modular BT design that enables a robot to adapt to environment uncertainties by pairing reactive condition monitoring with adaptive strategy selection, and learning from past experience to predict the optimal strategy given the task state. Our test robot used the adaptive BT to safely and robustly complete a diverse range of complex effort-intensive manipulation tasks commonly found in industrial and domestic settings, despite using out-the-box controllers, and with reduced time taken over the baseline BT design for all experiments where multiple strategies were available.
The adaptive BT can be readily added to existing BT implementations without major changes, enhancing performance while remaining interpretable and without significantly changing the system architecture. Future work will investigate algorithmic and learning-based synthesis of new strategies and conditions for the adaptive BT to enable a robot to generalize to new tasks, including synthesis from more semantic, conceptual descriptions of each strategy, or autonomous discovery of new strategies. 



\appendices
\section{Computing The Feasible Set of Strategies}
\label{appendix:feasible_set_of_strategies}
When manipulating articulated devices, all feasible strategies for task state $s$ are such that the robot arm joint efforts are not exceeded when using those strategies in the window $\{s':s_{L} \leq s'-s \leq s_{U}\}$ around $s$. Thus, $\mathcal{A}_{\mathrm{safe}}$ is given by
\begin{align*}
    \mathcal{A}_{\mathrm{safe}}(s) = \{& a \in \mathcal{A}_{\mathrm{task}} : |T_{\mathrm{est},i}(a,s')| \leq |T_{\mathrm{lim},i}(a)| \; \\
    & \forall i \in [1..N_{\mathrm{DOF}}], s_{L} \leq s'-s \leq s_{U} \},
\end{align*}
where $T_{\mathrm{est},i}(a,s)$ are the estimated efforts for each arm joint $i$ at task state $s$ using strategy $a$, recorded from previous manipulation attempts, and $T_{\mathrm{lim},i}(a)$ are the maximum safe efforts for each joint under strategy $a$.
$T_{\mathrm{est},i}$ are recorded from the joint efforts during each manipulation attempt. For valve manipulation, the task state $s$ is the total rotation angle $\theta$ away from being tight, with initial estimate 0 radians for loosening tasks. For tightening tasks, initial $\theta$ is estimated to be a large default value if unknown, but once the valve is tight, all $\theta$ values in the new batch of manipulation data are shifted such that the final entry is 0 radians. For button manipulation, the task has a (known) binary on/off state.

When grocery packing, the speed factor is computed from the additional effort that the weight of the object induces in the arm joints. Accordingly, $\mathcal{A}_{\mathrm{safe}}$ is given by
\begin{gather*}
    \mathcal{A}_{\mathrm{safe}}(s) = \{ a \in \mathbb{R}^{+} : a_{L} \leq a \leq a_{H}(s) \}\\
    \text{where } a_{H}(s)= \min \bigl\{a_{L}, 1 - \sum_{i} \alpha_{i} |T_{\mathrm{obj},i}(s) - T_{\mathrm{pre},i}|\bigr\},
\end{gather*}
where $T_{\mathrm{pre},i}$ and $T_{\mathrm{obj},i}(s)$ are the efforts recorded immediately before and after lifting the object respectively, $a_{L}$ is a minimum speed factor and $\alpha_{i} \geq 0$ are pre-tuned parameters. State $s$ is the object weight, but is not directly measured.


\bibliographystyle{IEEEtran}
\bibliography{IEEEabrv,references}

\end{document}